\documentclass{article}
\usepackage[utf8]{inputenc}
\usepackage{spconf,amsmath,amssymb,graphicx,booktabs}
\usepackage{microtype,xurl}
\usepackage[hidelinks]{hyperref}
\title{Human-1: Full-Duplex Spoken Dialogue Modeling for Hindi\\
       from Real-World Two-Channel Conversations}

\name{Bhaskar Singh, Manas Dhir, Shobhit Banga, Pranav Sharma}
\address{VoiceArena\\}

\begin{document}
\ninept
\maketitle

\begin{abstract}
Full-duplex spoken dialogue models capture interruptions, overlaps and
backchannels, yet exist almost exclusively for English. We present Human-1,
the first full-duplex spoken dialogue model for Hindi. Starting from Moshi,
whose tokenizer does not cover Devanagari, we train a Hindi tokenizer,
reinitialize vocabulary-dependent parameters while keeping the pre-trained
audio pathway and a frozen codec, and train in two stages on 26{,}000 hours of
spontaneous two-channel Hindi conversations from 14{,}695 speakers. Generated
speech attains a naturalness score of 4.10 versus 4.55 for human speech, with
66.9\% of paired comparisons ending in a tie, and its gap and pause statistics
closely match real conversations. On the IndicFDB benchmark, Human-1 produces
the most human-like backchannel timing of all evaluated systems, including
commercial ones. These results show that native two-channel conversational
data is sufficient to transfer full-duplex behavior to a new language and
script.
\end{abstract}


% ============================================================
\section{Introduction}
\label{sec:intro}
% ============================================================

Human conversation is inherently full-duplex: speakers overlap, backchannel
and interrupt one another, behaviors that half-duplex systems, which wait for
an end-of-turn before responding, cannot reproduce~\cite{nguyen2023dgslm,
wang2024fullduplex, ma2024listen}. Recent end-to-end speech dialogue models
address this by jointly modeling the user and system streams in
parallel~\cite{defossez2024moshi, veluri2024beyond, zhang2024omniflatten,
yu2024salmonn}.

This progress has been concentrated almost entirely on English. Hindi has more
than 600 million speakers, and its conversations feature frequent backchannels,
pervasive Hindi--English code-mixing, and turn-taking norms that differ from
those of English~\cite{hayashi1988simultaneous, stubbe1998listening}, yet no
full-duplex model exists for it. Adapting an existing duplex architecture to
Hindi raises three challenges.
(i)~\emph{Tokenization:} existing duplex models use English-centric text
vocabularies; Moshi's tokenizer does not cover Devanagari, so Hindi text cannot
be represented in its time-aligned text stream at all.
(ii)~\emph{Partial re-training:} replacing the tokenizer invalidates every
vocabulary-dependent parameter, so adaptation becomes partial pre-training
rather than conventional fine-tuning.
(iii)~\emph{Data:} duplex models must learn from two-channel recordings of
natural interaction, and resources at this scale are, to our knowledge,
unavailable for any Indian language.

We address all three and make the following contributions:
\begin{itemize}
\itemsep0em
\item A 26{,}000-hour corpus of spontaneous two-party Hindi conversations from
  14{,}695 speakers, recorded natively on separate channels, which to our
  knowledge is the largest purpose-built conversational speech corpus for an
  Indian language.
\item A language-adaptation recipe for Moshi: tokenizer replacement with
  targeted reinitialization, a frozen codec, and two-stage training, together
  with an analysis of its training dynamics.
\item Human-1, the first full-duplex spoken dialogue model for Hindi, evaluated
  with codec, perplexity, turn-taking and large-scale human evaluation, and
  benchmarked against commercial voice assistants on IndicFDB.
\end{itemize}

% ============================================================
\section{Related Work}
\label{sec:related}
% ============================================================

dGSLM~\cite{nguyen2023dgslm} was among the first models to generate two-channel
spoken dialogue jointly. Moshi~\cite{defossez2024moshi} made real-time
full-duplex dialogue practical by combining a 7B text LLM (Helium), a streaming
neural codec (Mimi) in the lineage of~\cite{zeghidour2021soundstream,
defossez2023encodec}, and a hierarchical RQ-Transformer.
SyncLLM~\cite{veluri2024beyond}, OmniFlatten~\cite{zhang2024omniflatten} and
SALMONN-omni~\cite{yu2024salmonn} explore alternative duplex formulations. All
of these systems target English or, in some cases, Chinese. Multilingual
speech--text models such as SpeechGPT~\cite{zhang2023speechgpt} and
AudioPaLM~\cite{rubenstein2023audiopalm} operate in a turn-based setting, and
cross-lingual transfer has mainly been studied for ASR~\cite{conneau2020xlsr}
and TTS~\cite{zhang2023vallex}. Adapting a full-duplex model to a new language,
especially one with a different script, has not been studied, and to our
knowledge no full-duplex system exists for any Indian language.

% ============================================================
\section{Hindi Conversational Corpus}
\label{sec:data}
% ============================================================

\textbf{Collection.} We collected 26{,}000 hours of unscripted two-party Hindi
conversations on diverse topics from 14{,}695 paid participants
(Table~\ref{tab:data}). Each speaker was recorded on a separate channel, which
yields true stereo dialogue. Moshi instead builds pseudo-stereo training data by
diarizing monophonic recordings~\cite{defossez2024moshi}; native channel
separation avoids the diarization errors and overlap artifacts of that
approach. Since turn-taking, backchannels and overlaps occur only in real
interaction, we deliberately avoided read speech, broadcast and monologue
sources. The corpus is part of an ongoing effort to collect one million hours
of spontaneous Hindi conversation.

\textbf{Quality control and alignment.} Trained annotators screened each
recording for clarity, speaker balance, transcription accuracy and
conversational naturalness, and recordings below threshold were discarded.
Transcripts were produced by a combination of manual and automatic
transcription, and word-level timestamps were obtained with
WhisperX~\cite{bain2023whisperx}. Text tokens are placed at their aligned
12.5\,Hz frames, and PAD tokens fill frames without text. Each training
sample spans 2{,}048 frames ($\approx$2.7\,min). The PAD ratio is
$\approx$75\%, higher than the $\approx$65\% reported for
English~\cite{defossez2024moshi}. We hypothesize that this reflects the
Hindi-specific vocabulary, which needs fewer text tokens per second of speech.

\textbf{Fine-tuning and test splits.} From the corpus we manually curate
1{,}000 hours with clear articulation, low background noise, balanced
participation and natural prosody. About 10 hours are held out for validation
and checkpoint selection, and the remaining $\approx$990 hours are used for
fine-tuning. All evaluations use a separately recorded set of conversations,
disjoint from the training corpus.

\begin{table}[t]
\centering
\caption{Hindi conversational corpus.}
\label{tab:data}
\small
\begin{tabular}{@{}ll@{}}
\toprule
Total duration & 26{,}000 h (1{,}000 h curated subset) \\
Unique speakers & 14{,}695 \\
Style & Spontaneous, unscripted, two-party \\
Channels & Stereo, one speaker per channel \\
Quality control & Annotator screening + manual checks \\
\bottomrule
\end{tabular}
\vspace{-2mm}
\end{table}

% ============================================================
\section{Method}
\label{sec:method}
% ============================================================

\subsection{Base architecture}
Moshi~\cite{defossez2024moshi} has two components. \textbf{Mimi} encodes
24\,kHz audio into 8 residual codebooks at 12.5\,Hz (1.1\,kbps); the first
codebook is distilled to carry semantic content and the remaining seven carry
acoustic detail. The \textbf{RQ-Transformer} consists of a 7B-parameter
\emph{Temporal Transformer}, which at each step $s$ consumes 17 parallel streams
(1 text, 8 system audio, 8 user audio) and outputs a context vector $z_s$ from
which a text token is predicted through a linear head, and a small
\emph{Depth Transformer}, which autoregressively predicts the 16 audio tokens
of step $s$ conditioned on $z_s$. Acoustic tokens are delayed by one step
relative to semantic tokens to stabilize generation.

We keep Mimi \textbf{frozen} throughout, because it reconstructs Hindi with
adequate quality without retraining (Sec.~\ref{sec:codec}). Only the
RQ-Transformer is trained, which substantially reduces cost.

\subsection{Tokenizer replacement and reinitialization}
\label{sec:tokenizer}
Moshi predicts a time-aligned text stream (its ``inner monologue'') alongside
audio, so it can only be trained on a language its text tokenizer can
represent. Moshi's SentencePiece~\cite{kudo2018sentencepiece} tokenizer was
trained for English and does not cover the Devanagari script, which makes
replacing it a prerequisite rather than an optimization. We train a
32{,}000-token SentencePiece model on a large Hindi text corpus. The
vocabulary size is unchanged, but the token identities change, so we
reinitialize every vocabulary-dependent parameter: the text embedding tables of
both the Temporal and Depth Transformers and the output text head. All other
weights, including the audio embeddings and the transformer blocks, are kept.
As a result, text representations are learned from scratch while the
pre-trained audio pathway is preserved, which makes the first stage
\emph{partial pre-training} rather than fine-tuning.

\subsection{Two-stage training}
\label{sec:training}
\textbf{Stage 1 (pre-training).} Because the text embeddings are freshly
initialized, we use Moshi's original pre-training learning rate,
$3{\times}10^{-5}$, instead of the smaller rates typical of adaptation, with
AdamW~\cite{loshchilov2019adamw} ($\beta_1{=}0.9$, $\beta_2{=}0.95$,
$\epsilon{=}10^{-5}$, weight decay 0.1). Following~\cite{defossez2024moshi},
the loss on PAD tokens is down-weighted by 50\% and semantic and acoustic
audio tokens are weighted 100:1. We train on 8$\times$H100 (80\,GB) GPUs in
bf16 with an effective batch of 64 samples (per-device batch 4, gradient
accumulation 2) for one epoch ($\approx$10k steps, $\approx$13\,h).

\textbf{Stage 2 (fine-tuning).} Starting from the Stage-1 checkpoint, we
fine-tune on the $\approx$990-hour curated set with an effective batch of 16,
50 warm-up steps and component-specific learning rates: $2{\times}10^{-6}$ for
the Temporal Transformer and $4{\times}10^{-6}$ for the Depth Transformer. The
higher Depth-Transformer rate allows faster adaptation of acoustic prediction
while limiting drift in the newly learned text representations. We validate
and checkpoint every 802 steps.

% ============================================================
\section{Experiments and Results}
\label{sec:results}
% ============================================================

\subsection{Training dynamics}
\label{sec:dynamics}

\begin{figure}[t]
\centering
\includegraphics[width=\columnwidth]{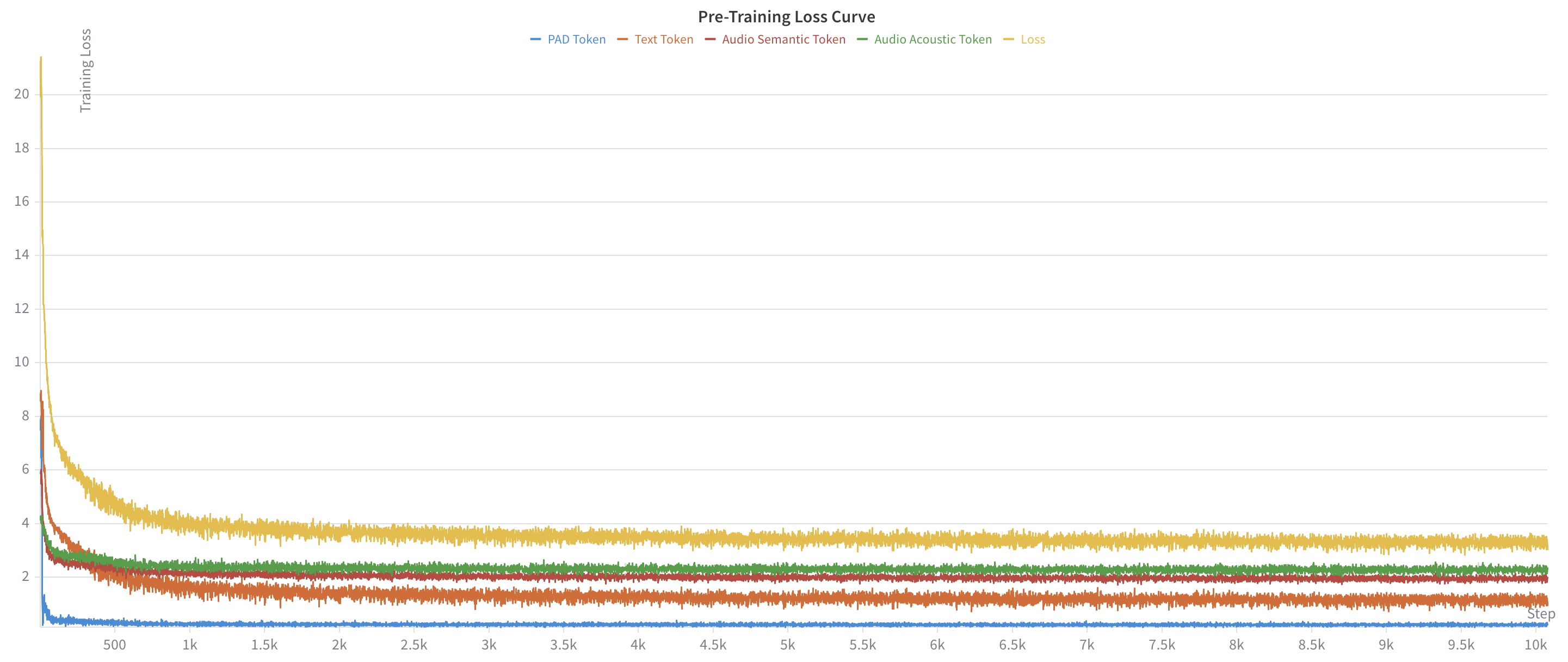}
\vspace{-6mm}
\caption{Stage-1 training loss on 26{,}000\,h of Hindi conversation. All
components converge within 2k--4k steps.}
\label{fig:pretrain}
\vspace{-3mm}
\end{figure}

\textbf{Pre-training.} All loss components drop sharply during the first
$\sim$2k steps (Fig.~\ref{fig:pretrain}), and the total loss falls from
$\approx$20 at initialization to $\approx$3.5, where it plateaus after
$\sim$4k steps, i.e., before half an epoch. At convergence, text accuracy on
non-PAD tokens is $\approx$70\% and audio accuracy is 48--53\% across
codebooks. Even though the text embeddings and head start from random
initialization, the text loss converges almost as fast as the audio losses,
which indicates that the retained transformer weights transfer across scripts.

\begin{figure}[t]
\centering
\includegraphics[width=\columnwidth]{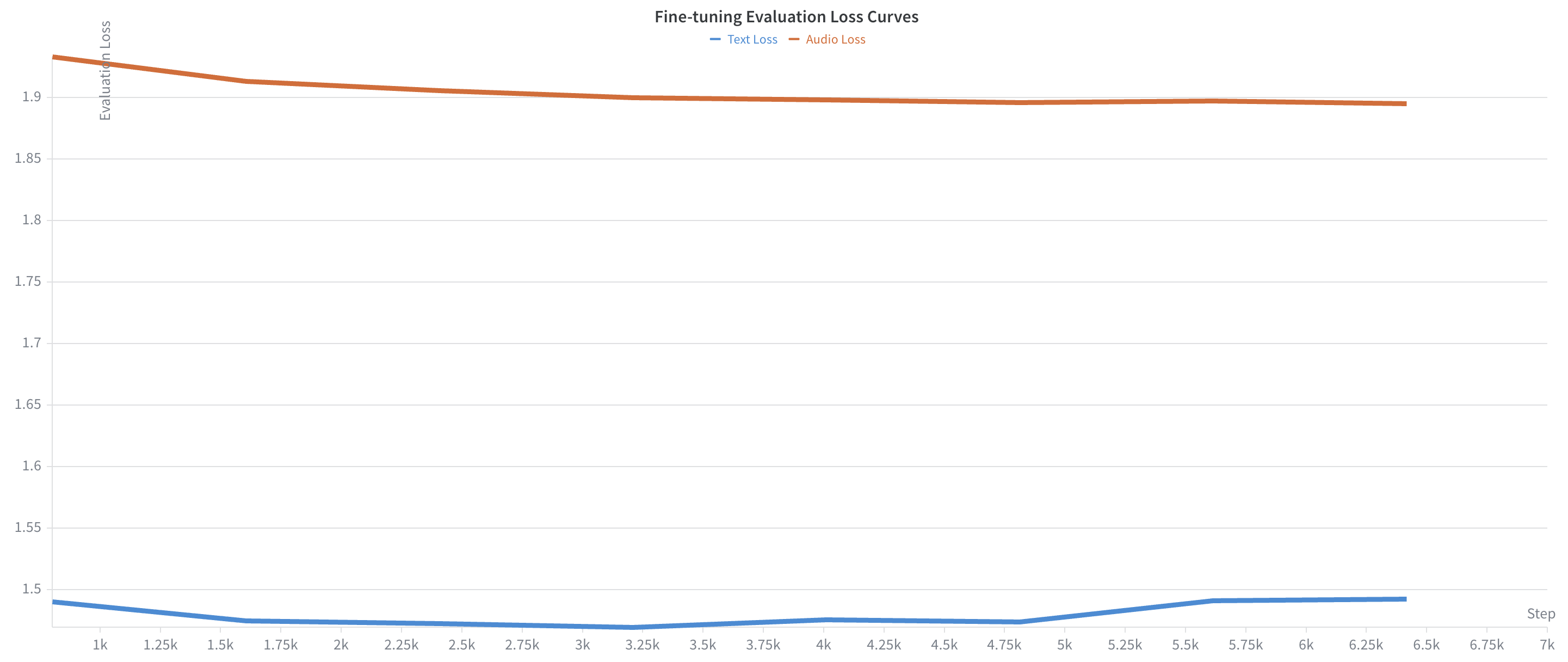}
\vspace{-6mm}
\caption{Stage-2 validation loss. Text loss overfits beyond $\sim$4.8k steps
while audio loss improves marginally. The selected checkpoint is step 4{,}812.}
\label{fig:finetune}
\vspace{-3mm}
\end{figure}

\textbf{Fine-tuning.} The text and audio losses behave differently on the
validation set (Fig.~\ref{fig:finetune}). The text validation loss flattens
after $\sim$1.6k steps and rises steadily beyond $\sim$4.8k steps, a clear sign
of overfitting to the curated subset. The audio validation loss continues to
decrease slowly with diminishing returns. We select the checkpoint at step
4{,}812, which minimizes total validation loss (3.370; text 1.474, audio 1.896).
Training loss continues to decrease well past this point, so validation-based
early stopping is essential when fine-tuning on limited curated data.

\subsection{Evaluation protocol}
We follow the prompted dialogue continuation protocol~\cite{nguyen2023dgslm,
defossez2024moshi}. Each held-out dialogue is split into 30\,s segments: the
first 10\,s of both channels serve as the prompt, and the model generates the
next 20\,s at sampling temperature $\tau \in \{0.8, 0.9, 1.0\}$.

\subsection{Codec fidelity on Hindi}
\label{sec:codec}
To test whether Mimi, trained predominantly on English, generalizes to Hindi, we
compare 654 twenty-second test segments with their Mimi re-syntheses. STOI
of $0.878{\pm}0.027$~\cite{taal2011stoi} indicates that intelligibility is
largely preserved. PESQ of $2.55{\pm}0.37$~\cite{rix2001pesq} is moderate,
which is expected at 1.1\,kbps for a codec designed for low-latency streaming
rather than maximal fidelity. On this basis we freeze Mimi.

\subsection{Linguistic quality and turn-taking}
We transcribe the generated speech with Whisper-large-v3~\cite{radford2023whisper}
and score the transcripts with Sarvam-1 (2B)~\cite{sarvam1}, a Hindi-capable
LM, reporting perplexity (PPL) as in~\cite{nguyen2023dgslm,
defossez2024moshi}. For turn-taking we apply energy-based VAD independently to
each channel and compute the cumulative duration (s/min) of inter-pausal units
(IPU), pauses (within-speaker silences), gaps (between-speaker silences) and
overlaps~\cite{nguyen2023dgslm}.

\begin{table}[t]
\centering
\caption{Perplexity of generated speech (Whisper-v3 transcripts scored by
Sarvam-1) and turn-taking statistics (s/min).}
\label{tab:auto}
\small
\setlength{\tabcolsep}{4.5pt}
\begin{tabular}{@{}lcccccc@{}}
\toprule
 & $\tau$ & PPL$\downarrow$ & IPU & Pause & Gap & Overlap \\
\midrule
Ground truth & -- & 237.1 & 35.30 & 10.49 & 8.51 & 3.03 \\
\midrule
Human-1 & 0.8 & \textbf{356.9} & 23.12 & 9.16 & 6.77 & 1.67 \\
 & 0.9 & 467.1 & 29.14 & \textbf{9.24} & \textbf{8.54} & \textbf{4.30} \\
 & 1.0 & 640.6 & \textbf{38.90} & 11.67 & 8.10 & 9.68 \\
\bottomrule
\end{tabular}
\vspace{-2mm}
\end{table}

Table~\ref{tab:auto} shows a trade-off controlled by temperature. PPL is lowest
at $\tau{=}0.8$ and rises with $\tau$ as sampling becomes more diverse. The
turn-taking statistics, however, are closest to ground truth at $\tau{=}0.9$:
the gap duration nearly matches the reference (8.54 vs.\ 8.51\,s/min) and the
pause duration is comparable (9.24 vs.\ 10.49). At $\tau{=}0.8$ the model is
overly conservative, with less speech and little overlap, whereas at
$\tau{=}1.0$ overlap is three times the reference (9.68 vs.\ 3.03). The
ground-truth PPL of 237.1 is itself high, which reflects ASR errors on
spontaneous, code-mixed speech and the domain mismatch between written-text LMs
and conversational transcripts. PPL is therefore best interpreted relative to
this reference.

\subsection{Human evaluation}
In total, 130 native Hindi speakers completed 2{,}125 paired comparisons
between human and model continuations of the same prompt. Raters scored
naturalness and clarity on 5-point scales, stated a preference, and answered
binary rubric questions on conversational quality (Table~\ref{tab:human}).

\begin{table}[t]
\centering
\caption{Human evaluation (130 raters, 2{,}125 paired comparisons). Rubric
pass rates are rounded.}
\label{tab:human}
\small
\begin{tabular}{@{}lcc@{}}
\toprule
\textbf{Perceptual (1--5)} & Human & Human-1 \\
\midrule
Naturalness & 4.55 & 4.10 \\
Clarity & 4.05 & 3.04 \\
\midrule
Preferred (tie: 66.9\%) & 30.0\% & 3.1\% \\
\midrule
\textbf{Rubric pass rate (Human-1)} & \multicolumn{2}{c}{} \\
\midrule
Human-like interaction & \multicolumn{2}{c}{85\%} \\
Appropriateness (follows prompt) & \multicolumn{2}{c}{53\%} \\
Completion (complete reply) & \multicolumn{2}{c}{42\%} \\
\bottomrule
\end{tabular}
\vspace{-2mm}
\end{table}

Human-1 reaches a naturalness score of 4.10, against 4.55 for human speech, and
two thirds of the pairwise comparisons (66.9\%) are ties, so listeners often
cannot distinguish its speech from real conversation. Clarity (3.04 vs.\ 4.05)
is weaker, which is consistent with the moderate PESQ of the frozen codec.
The rubric results locate the main limitation in content rather than form:
85\% of responses are judged human-like, but only 53\% follow the prompt
context appropriately and 42\% form complete replies. Semantic coherence over
longer contexts is therefore the main direction for improvement.

\subsection{Benchmark comparison on IndicFDB}
\label{sec:indicfdb}
We further evaluate Human-1 on IndicFDB~\cite{indicfdb}, a concurrent
benchmark to which some of the present authors contributed, which extends the
Full-Duplex-Bench protocol~\cite{lin2025fdb} to Indian languages and measures
four interactive behaviors: \emph{pause handling} (not taking the floor during
a user's mid-utterance pause), \emph{turn-taking} (responding when the user
finishes), \emph{backchanneling} (success rate, Jensen--Shannon distance
(JSD) between agent and human backchannel timing, and backchannel frequency)
and \emph{user interruption} (yielding the floor, latency, and a 0--5 LLM
rating of the relevance and quality of the post-interruption response,
obtained after transcription and translation to English). Table~\ref{tab:indicfdb}
reports the Hindi results. Apart from Human-1, all evaluated systems are
proprietary production voice assistants.

\begin{table*}[t]
\centering
\caption{Hindi results on IndicFDB~\cite{indicfdb}. S.R.: success rate (\%);
Lat.: latency (s); JSD: Jensen--Shannon distance; Freq.: backchannels/s
(closer to Human is better); Rating: interruption response rating (0--5). All systems except
Human-1 are proprietary.}
\label{tab:indicfdb}
\small
\begin{tabular}{@{}lccccccccc@{}}
\toprule
 & Pause handling & \multicolumn{2}{c}{Turn taking} & \multicolumn{3}{c}{Agent backchanneling} & \multicolumn{3}{c}{User interruption} \\
\cmidrule(lr){2-2}\cmidrule(lr){3-4}\cmidrule(lr){5-7}\cmidrule(l){8-10}
Model & S.R.$\uparrow$ & S.R.$\uparrow$ & Lat.$\downarrow$ & S.R.$\uparrow$ & JSD$\downarrow$ & Freq.$\leftrightarrow$ & S.R.$\uparrow$ & Lat.$\downarrow$ & Rating$\uparrow$ \\
\midrule
Human      & 100.00 & 100.00 & 0.259 & 100.00 & 0.000 & 0.079 & -- & -- & -- \\
\midrule
GPT Live   & 61.59 & 91.06 & \textbf{0.181} & 15.32 & 0.822 & \textbf{0.109} & \textbf{100.00} & \textbf{0.322} & 4.857 \\
Grok       & 99.05 & 85.11 & 2.556 & 98.18 & 0.986 & 0.004 & 99.67 & 1.628 & 4.960 \\
Gemini 3.8 & \textbf{100.00} & \textbf{99.15} & 1.400 & 98.18 & 0.942 & 0.010 & \textbf{100.00} & 1.255 & 4.970 \\
Gemini 3.1 & \textbf{100.00} & 98.30 & 2.128 & \textbf{100.00} & 0.990 & 0.003 & \textbf{100.00} & 1.678 & \textbf{4.973} \\
\midrule
Human-1 (ours) & 72.06 & 89.36 & 2.021 & 45.97 & \textbf{0.795} & 0.110 & 95.67 & 1.573 & 0.564 \\
\bottomrule
\end{tabular}
\vspace{-3mm}
\end{table*}

Human-1 achieves the lowest backchannel JSD of all systems (0.795), i.e., its
backchannel timing is the closest to human behavior, and its backchannel
frequency (0.110 vs.\ 0.079 for humans) is on par with the best system
(0.109), while its backchannel success rate is three times that of GPT Live
(45.97\% vs.\ 15.32\%). The commercial systems other than GPT Live either almost never
backchannel (frequency $\le$0.010) or do so with timing far from human
(JSD $\ge$0.94), despite high success rates. This supports our central claim
that training on native two-channel conversations teaches listener behavior
that is otherwise difficult to obtain. In turn-taking success Human-1 is close
to GPT Live (89.36\% vs.\ 91.06\%) and ahead of Grok (85.11\%), and it
outperforms GPT Live in pause handling (72.06\% vs.\ 61.59\%). Human-1 also
yields the floor in 95.67\% of user interruptions. Its main weaknesses are
response latency ($\approx$2\,s) and the low rating of its post-interruption
responses (0.564). The same judging pipeline rates the other Hindi systems at 4.857--4.973, so
transcription and translation errors are unlikely to explain this gap
fully~\cite{indicfdb}. The low rating is consistent with the appropriateness and
completion rubrics in Table~\ref{tab:human}: Human-1 handles the timing of
interaction well, but the semantic content of its replies lags behind systems
built on large proprietary text LLMs. For reference, the original English
Moshi, which cannot operate in Hindi, obtains 57.46\% pause handling, 71.06\%
turn-taking and 3.12\% backchannel success on the Indian-English split of the
same benchmark~\cite{indicfdb}; Human-1 exceeds all three in Hindi.

% ============================================================
\section{Conclusion}
\label{sec:conclusion}
% ============================================================

We presented Human-1, a full-duplex spoken dialogue model for Hindi obtained by
adapting Moshi through tokenizer replacement, targeted reinitialization and
two-stage training on 26{,}000 hours of native two-channel conversations.
Mimi transfers to Hindi without retraining, so adaptation can be confined to
the RQ-Transformer. The model learns realistic turn-taking directly from real
conversations, without synthetic dialogue data, and on IndicFDB its
backchannel timing is closer to human behavior than that of any commercial
system evaluated. We also report training
dynamics that can guide adaptation of Moshi to other languages. Moshi was
trained on about 7M hours of speech, whereas Human-1 reaches natural full-duplex
behavior with 26k hours. Scaling quality-controlled conversational data
therefore appears to be the most promising route to fluent and coherent duplex
dialogue for Hindi and other Indian languages.

\bibliographystyle{IEEEbib}
\bibliography{refs}

@article{nguyen2023dgslm,
  title   = {Generative Spoken Dialogue Language Modeling},
  author  = {Nguyen, Tu Anh and Kharitonov, Eugene and Copet, Jade and Adi, Yossi and
             Hsu, Wei-Ning and Elkahky, Ali and Tomasello, Paden and Algayres, Robin and
             Sagot, Beno{\^\i}t and Mohamed, Abdelrahman and Dupoux, Emmanuel},
  journal = {Trans. Assoc. Comput. Linguist.},
  volume  = {11},
  pages   = {250--266},
  year    = {2023}
}

@inproceedings{wang2024fullduplex,
  title     = {A Full-Duplex Speech Dialogue Scheme Based on Large Language Model},
  author    = {Wang, Peng and Lu, Songshuo and Tang, Yaohua and Yan, Sijie and
               Xia, Wei and Xiong, Yuanjun},
  booktitle = {Proc. Adv. Neural Inf. Process. Syst. (NeurIPS)},
  year      = {2024}
}

@article{ma2024listen,
  title   = {Language Model Can Listen While Speaking},
  author  = {Ma, Ziyang and Song, Yakun and Du, Chenpeng and Cong, Jian and
             Chen, Zhuo and Wang, Yuping and Wang, Yuxuan and Chen, Xie},
  journal = {arXiv preprint arXiv:2408.02622},
  year    = {2024}
}

@article{defossez2024moshi,
  title   = {Moshi: A Speech-Text Foundation Model for Real-Time Dialogue},
  author  = {D{\'e}fossez, Alexandre and Mazar{\'e}, Laurent and Orsini, Manu and
             Royer, Am{\'e}lie and P{\'e}rez, Patrick and J{\'e}gou, Herv{\'e} and
             Grave, Edouard and Zeghidour, Neil},
  journal = {arXiv preprint arXiv:2410.00037},
  year    = {2024}
}

@inproceedings{veluri2024beyond,
  title     = {Beyond Turn-Based Interfaces: Synchronous {LLM}s as Full-Duplex Dialogue Agents},
  author    = {Veluri, Bandhav and Peloquin, Benjamin N. and Yu, Bokai and
               Gong, Hongyu and Gollakota, Shyamnath},
  booktitle = {Proc. Conf. Empirical Methods Nat. Lang. Process. (EMNLP)},
  pages     = {21390--21402},
  year      = {2024}
}

@article{zhang2024omniflatten,
  title   = {{OmniFlatten}: An End-to-End {GPT} Model for Seamless Voice Conversation},
  author  = {Zhang, Qinglin and Cheng, Luyao and Deng, Chong and Chen, Qian and
             Wang, Wen and Zheng, Siqi and Liu, Jiaqing and Yu, Hai and Tan, Chaohong},
  journal = {arXiv preprint arXiv:2410.17799},
  year    = {2024}
}

@article{yu2024salmonn,
  title   = {{SALMONN-omni}: A Codec-Free {LLM} for Full-Duplex Speech Understanding and Generation},
  author  = {Yu, Wenyi and Wang, Siyin and Yang, Xiaoyu and Chen, Xianzhao and
             Tian, Xiaohai and Zhang, Jun and Sun, Guangzhi and Lu, Lu and
             Wang, Yuxuan and Zhang, Chao},
  journal = {arXiv preprint arXiv:2411.18138},
  year    = {2024}
}

@article{hayashi1988simultaneous,
  title   = {Simultaneous Talk---From the Perspective of Floor Management of {E}nglish and {J}apanese Speakers},
  author  = {Hayashi, Reiko},
  journal = {World Englishes},
  volume  = {7},
  number  = {3},
  pages   = {269--288},
  year    = {1988}
}

@article{stubbe1998listening,
  title   = {Are You Listening? {C}ultural Influences on the Use of Supportive Verbal Feedback in Conversation},
  author  = {Stubbe, Maria},
  journal = {J. Pragmatics},
  volume  = {29},
  number  = {3},
  pages   = {257--289},
  year    = {1998}
}

@inproceedings{zhang2023speechgpt,
  title     = {{SpeechGPT}: Empowering Large Language Models with Intrinsic Cross-Modal Conversational Abilities},
  author    = {Zhang, Dong and Li, Shimin and Zhang, Xin and Zhan, Jun and
               Wang, Pengyu and Zhou, Yaqian and Qiu, Xipeng},
  booktitle = {Findings Assoc. Comput. Linguist. (EMNLP)},
  pages     = {15757--15773},
  year      = {2023}
}

@article{rubenstein2023audiopalm,
  title   = {{AudioPaLM}: A Large Language Model That Can Speak and Listen},
  author  = {Rubenstein, Paul K. and Asawaroengchai, Chulayuth and Nguyen, Duc Dung and others},
  journal = {arXiv preprint arXiv:2306.12925},
  year    = {2023}
}

@inproceedings{conneau2020xlsr,
  title     = {Unsupervised Cross-Lingual Representation Learning for Speech Recognition},
  author    = {Conneau, Alexis and Baevski, Alexei and Collobert, Ronan and
               Mohamed, Abdelrahman and Auli, Michael},
  booktitle = {Proc. Interspeech},
  pages     = {2426--2430},
  year      = {2021}
}

@article{zhang2023vallex,
  title   = {Speak Foreign Languages with Your Own Voice: Cross-Lingual Neural Codec Language Modeling},
  author  = {Zhang, Ziqiang and Zhou, Long and Wang, Chengyi and others},
  journal = {arXiv preprint arXiv:2303.03926},
  year    = {2023}
}

@article{defossez2023encodec,
  title   = {High Fidelity Neural Audio Compression},
  author  = {D{\'e}fossez, Alexandre and Copet, Jade and Synnaeve, Gabriel and Adi, Yossi},
  journal = {Trans. Mach. Learn. Res.},
  year    = {2023}
}

@article{zeghidour2021soundstream,
  title   = {{SoundStream}: An End-to-End Neural Audio Codec},
  author  = {Zeghidour, Neil and Luebs, Alejandro and Omran, Ahmed and
             Skoglund, Jan and Tagliasacchi, Marco},
  journal = {IEEE/ACM Trans. Audio, Speech, Lang. Process.},
  volume  = {30},
  pages   = {495--507},
  year    = {2022}
}

@inproceedings{kudo2018sentencepiece,
  title     = {{SentencePiece}: A Simple and Language Independent Subword Tokenizer and Detokenizer for Neural Text Processing},
  author    = {Kudo, Taku and Richardson, John},
  booktitle = {Proc. Conf. Empirical Methods Nat. Lang. Process. (EMNLP): Syst. Demonstrations},
  pages     = {66--71},
  year      = {2018}
}

@inproceedings{bain2023whisperx,
  title     = {{WhisperX}: Time-Accurate Speech Transcription of Long-Form Audio},
  author    = {Bain, Max and Huh, Jaesung and Han, Tengda and Zisserman, Andrew},
  booktitle = {Proc. Interspeech},
  pages     = {4489--4493},
  year      = {2023}
}

@inproceedings{radford2023whisper,
  title     = {Robust Speech Recognition via Large-Scale Weak Supervision},
  author    = {Radford, Alec and Kim, Jong Wook and Xu, Tao and Brockman, Greg and
               McLeavey, Christine and Sutskever, Ilya},
  booktitle = {Proc. Int. Conf. Mach. Learn. (ICML)},
  pages     = {28492--28518},
  year      = {2023}
}

@misc{sarvam1,
  title        = {{Sarvam-1}: A 2{B} Parameter Language Model Optimized for {I}ndian Languages},
  author       = {{Sarvam AI}},
  howpublished = {\url{https://huggingface.co/sarvamai/sarvam-1}},
  year         = {2024}
}

@inproceedings{rix2001pesq,
  title     = {Perceptual Evaluation of Speech Quality ({PESQ})---A New Method for Speech Quality Assessment of Telephone Networks and Codecs},
  author    = {Rix, Antony W. and Beerends, John G. and Hollier, Michael P. and Hekstra, Andries P.},
  booktitle = {Proc. IEEE Int. Conf. Acoust., Speech, Signal Process. (ICASSP)},
  pages     = {749--752},
  year      = {2001}
}

@article{taal2011stoi,
  title   = {An Algorithm for Intelligibility Prediction of Time--Frequency Weighted Noisy Speech},
  author  = {Taal, Cees H. and Hendriks, Richard C. and Heusdens, Richard and Jensen, Jesper},
  journal = {IEEE Trans. Audio, Speech, Lang. Process.},
  volume  = {19},
  number  = {7},
  pages   = {2125--2136},
  year    = {2011}
}

@inproceedings{loshchilov2019adamw,
  title     = {Decoupled Weight Decay Regularization},
  author    = {Loshchilov, Ilya and Hutter, Frank},
  booktitle = {Proc. Int. Conf. Learn. Represent. (ICLR)},
  year      = {2019}
}

@article{lin2025fdb,
  title   = {{Full-Duplex-Bench}: A Benchmark to Evaluate Full-Duplex Spoken Dialogue Models on Turn-Taking Capabilities},
  author  = {Lin, Guan-Ting and Lian, Jiachen and Li, Tingle and Wang, Qirui and
             Anumanchipalli, Gopala and Liu, Alexander H. and Lee, Hung-yi},
  journal = {arXiv preprint arXiv:2503.04721},
  year    = {2025}
}

@misc{indicfdb,
  title        = {{IndicFDB}: Benchmarking Full-Duplex Voice Agents across {I}ndian Languages},
  author       = {Roy, Rajarshi and Banga, Shobhit and Raiman, Jonathan and Paul, Supriya and
                  Singh, Bhaskar and Kaur, Manmeet and Jain, Sagar and Sidh, Hanuman and
                  Sharma, Pranav and Singh, Aditya and Pareek, Aaditya and Dhir, Manas and
                  Margolin, Adi and Agarwal, Niket and Catanzaro, Bryan},
  howpublished = {\url{https://3bk47sbkf7b.github.io/indicfdb/paper.pdf}},
  year         = {2026}
}

\end{document}